\documentclass[
]{ceurart}

\usepackage{listings}
\usepackage{latexsym}
\usepackage{amssymb}
\usepackage{amsmath}

\usepackage{amsfonts}
\usepackage{booktabs}
\usepackage{enumitem}
\usepackage{graphicx}
\usepackage{color}
\usepackage{comment}
\usepackage{hyperref}

\usepackage{mathtools}
\usepackage{multirow}
\usepackage{listings}
\usepackage{xcolor}
\usepackage{tabularx}
\usepackage{eurosym}
\usepackage{algorithm}
\usepackage{algpseudocode}
\usepackage{pifont}
\usepackage[switch]{lineno}

\newcolumntype{Y}{>{\centering\arraybackslash}X}

\DeclarePairedDelimiter\abs{\lvert}{\rvert}%

\newcommand{\R}{\mathbb{R}}

\newcommand{\odotC}{\odot_{\mathbb{C}}}
\newcommand{\cat}{\mathbin\Vert}
\newcommand{\calG}{\mathcal{G}}
\newcommand{\calR}{\mathcal{R}}
\newcommand{\calE}{\mathcal{E}}
\newcommand{\calT}{\mathcal{T}}
\newcommand{\calN}{\mathcal{N}}

\newcommand{\calNin}{\mathcal{N}_{\mathrm{in}}}

\newcommand{\RSPMM}{\mathrm{RSPMM}}

\newcommand{\rspmm}[4]{\left\langle #2 \middle| #3 \middle| #4 \right\rangle_{#1}}

\newcommand{\bfh}{\mathbf{h}}
\newcommand{\bfH}{\mathbf{H}}
\newcommand{\bfz}{\mathbf{z}}
\newcommand{\bfZ}{\mathbf{Z}}

\copyrightyear{2026}
\copyrightclause{Copyright for this paper by its authors. Use permitted under Creative Commons License Attribution 4.0 International (CC BY 4.0).}
\conference{ISWC 2026 Companion Volume, October 25--29, 2026, Bari, Italy}

\begin{document}

\copyrightyear{2022}
\copyrightclause{Copyright for this paper by its authors.
  Use permitted under Creative Commons License Attribution 4.0
  International (CC BY 4.0).}

\conference{ISWC'26: Posters and Demos Track,
  October 25--29, 2026, Bari, Italy}

\title{Scalable GNN-based Knowledge Graph Representation Learning with Efficient Message Passing}

\tnotemark[1]
\tnotetext[1]{This poster is accompanying our conference paper~\cite{mai2026rspmm}.}

\author[1,2]{Huu Tan Mai}[%
email=huutan.mai@de.bosch.com,
]
\cormark[1]

\author[1]{Cuong Xuan Chu}[%
email=cuongxuan.chu@de.bosch.com,
]

\author[2]{Heiko Paulheim}[%
email=heiko.paulheim@uni-mannheim.de,
]

\author[1]{Daria Stepanova}[%
email=daria.stepanova@de.bosch.com,
]
\cormark[1]

\address[1]{Bosch Center for Artificial Intelligence,
  Robert-Bosch-Campus 1, 71272 Renningen, Germany}
\address[2]{University of Mannheim,
  Schloss, 68131 Mannheim, Germany}

\cortext[1]{Corresponding author.}

\begin{abstract}
Graph neural networks (GNNs) excel at representation learning on Knowledge Graphs (KGs), achieving state-of-the-art performance on 
tasks like 
link prediction or entity classification. However, their high computational complexity, inherent to their user-defined message passing (MP) algorithm, still prohibits their widespread adoption, especially for large KGs. 
Current efforts to mitigate the 
scalability bottlenecks of GNNs on KGs, such as subgraph sampling, are often task- and model-specific, and do not reliably guarantee lossless (if applicable) runtime/space reductions.
To address this, we extend Relational Sparse Matrix Multiplication (RSPMM), originally designed to losslessly lower the space complexity of composition-based MP with pointwise composition functions, to support more expressive functions (e.g., $2\times2$ block-diagonal matrix multiplication, Givens rotation, circular correlation). Our method delivers significant task-independent reductions in runtime and space for current GNNs on KGs and facilitates efficient re-implementations of GNNs that maintain near state-of-the-art performance on challenging KG tasks, for a fraction of computational costs.
\end{abstract}

\maketitle

\section{Introduction}

Knowledge Graphs (KGs) (e.g., YAGO~\cite{Mahdisoltani2015YAGO3AK}, Wikidata~\cite{wikidata}) represent domain knowledge as multi-relational directed graphs of facts, encoded as labeled edges between entities. Graph representation learning~\cite{pytorch-biggraph,hamilton2018inductiverepresentationlearninglarge,relational-ml} has emerged as a powerful paradigm that enables machine learning on KGs for tasks such as KG completion (KGC)~\cite{transe,distmult,complex,nbfnet}, entity classification~\cite{wl-rdf,rdf2vec,rgcn} or alignment~\cite{ge2021largeeaaligningentitieslargescale,zhu2017iterative}. Two major examples of representation learning approaches are shallow KG embeddings (KGE)~\cite{transe,distmult,complex,rotate,nickel2015holographicembeddingsknowledgegraphs,dettmers2018convolutional2dknowledgegraph}, and Graph Neural Networks (GNN)~\cite{rgcn,compgcn,kracl,ragat,nbfnet}. Shallow KGEs directly embed entities and relations into a low-dimensional vector space and are the \textit{de facto} method for web-scale graphs~\cite{kepler} thanks to their simple design and dedicated software~\cite{zhu2019graphvite,DGL-KE}, but they are limited in terms of accuracy. In contrast, GNNs yield higher predictive accuracy by using message passing (MP) 
to learn expressive node and relation representations 
from neighborhoods, but they are computationally expensive.

\paragraph{Notations.} Given a KG $\calG = (\calE, \calR, \calT)$, $A = \left(\alpha_{jri}\right) \in \mathbb{R}^{\abs{\calE} \times \abs{\calR} \times \abs{\calE}}$ is an \emph{adjacency tensor} of $\mathcal{G}$ when it verifies $(j,r,i) \notin \calT \implies \alpha_{jri} = 0$ (non-existent edges have zero weight). Furthermore, we define
 \begin{align*}
 \calNin(i) = \left\{(j, r) \in \calE \times \calR : (j, r, i) \in \calT \right\}.
\end{align*}

\begin{figure}[t]
    \centering
    \includegraphics[width=\linewidth]{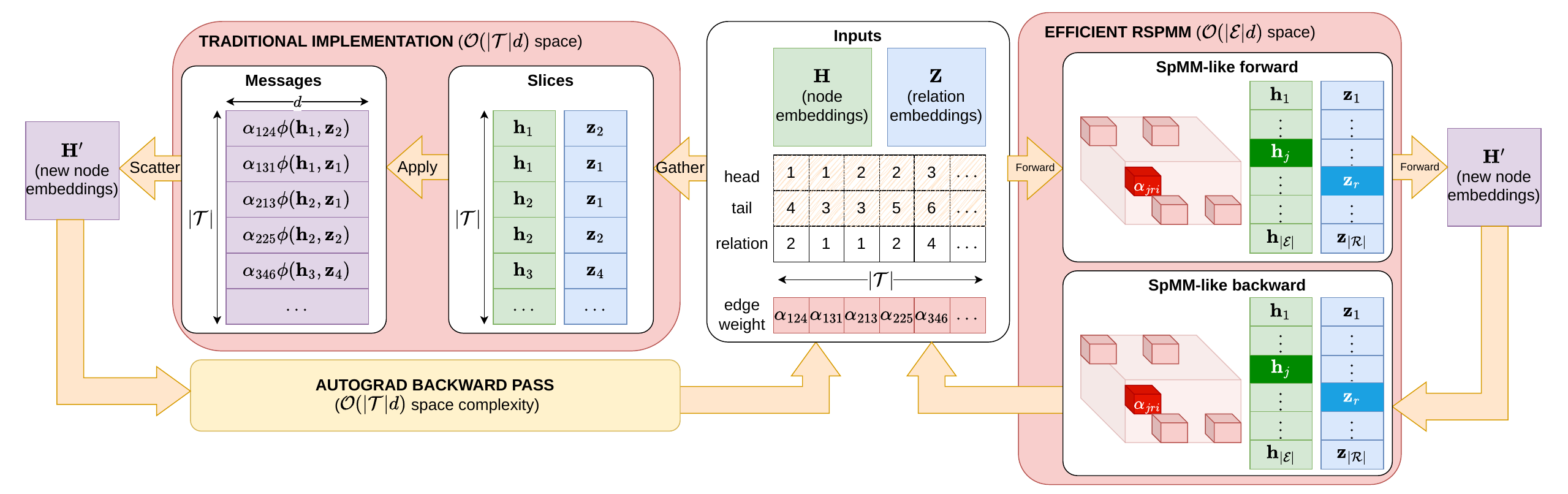}
    \caption{Example of traditional implementations  of composition-based MP (left), against efficient $\RSPMM_\phi$ implementations (right). The latter aim to fuse all operations in lower space complexity, as in practice, $\abs{\calE} \ll \abs{\calT}$.}
    \label{fig:rspmm}
\end{figure}

\paragraph{Problem.} 
For a KG $\mathcal{G} = (\calE, \calR, \calT)$ with entities $\calE$, relations $\calR$ and triples $\calT$, and an adjacency tensor $A = (\alpha_{jri}) \in \mathbb{R}^{\abs{\calE} \times \abs{\calR} \times \abs{\calE}}$, we consider the main building block of the popular composition-based MP method (e.g.,~\cite{compgcn,ragat,kracl,nbfnet,liang2023hyper,chen2021rgatrelationalgraphattention,adaprop}). The node update equation is typically defined as follows:
\begin{align}\label{eq:composition-mp}
    \bfH'_i = \mathrm{UPD}\left(\bfH_{i}, \mathbf{M}_i\right),\quad \mathbf{M}_i = \bigoplus_{ (j, r) \in \calN_{\mathrm{in}}(i)} \mathbf{m}_{jri},\quad \mathbf{m}_{jri} = \alpha_{jri} \phi\left(\bfH_j, \mathbf{Z}_r \right),\
\end{align}
where $\bfH$ is the matrix of input entity representations, $\mathbf{Z}$ is the matrix of input relation representations and $\bfH'$ is the matrix of output entity representations, $\bigoplus$ is a permutation invariant aggregation and $\mathrm{UPD}$ is an update operation. Traditionally, computing $\bfH'$ from $\bfH$ and $\mathbf{Z}$ requires a gather-scatter methodology (e.g., codebases of~\cite{compgcn,ragat,adaprop}):  (1) \textbf{gather} $\bfH$ and $\mathbf{Z}$ rows, then compute $\mathbf{m}_{jri}$ for \textbf{each edge}, and (2) aggregate using a \textbf{scatter} function (e.g., \texttt{scatter\_add} in PyG~\cite{pyg}). If $\mathbf{m}_{jri} \in \R^d$, then the step (2) generates a tensor of size $\mathcal{O}(\abs{\calT}d)$ in memory (see Figure~\ref{fig:rspmm} for illustration). Unless specific simplifications or high-performance CUDA kernels are available, complex MP designs employ this abstraction, and the induced memory footprint is a major hurdle for scalable GNN-based graph learning.

\paragraph{State-of-the-Art and its Limitations.} While vocabulary reduction~\cite{nodepiece,recpiece,nbfnet,astarnet} and subgraph sampling~\cite{adaprop,astarnet,feeney2021relationmatterssamplingscalable,stargraph} have been explored to improve the parameter efficiency and scalability of relation-aware GNNs, they do not address the very 
costly MP methods. $\RSPMM_{\phi}, \phi \in \{ +, \odot \}$ (\textbf{R}elational \textbf{Sp}arse \textbf{M}atrix \textbf{M}ultiplication) was introduced~\cite{nbfnet} to compute $\mathbf{M}$ in $\mathcal{O}(\abs{\calE}d)$ space complexity, as a fused MP CUDA kernel that modifies GE-SpMM~\cite{ge-spmm}. Since $\abs{\calE} \ll \abs{\calT}$ holds in practice, this improvement is 
beneficial for the scalability of GNNs using $\RSPMM_{+ / \odot}$.
However, this option is not trivially available for GNNs with expressive messages 
(e.g., dense linear transformation, rotation, etc.).

\paragraph{Contributions.} To address the above scalability and genericity challenges, inspired by NBFNet~\cite{nbfnet}, we introduce novel and generic scalability optimizations of relational MP methods by generalizing the aforementioned RSPMM to be applicable to other works. Many existing GNN methods~\cite{compgcn,ragat,kracl,liang2023hyper,nbfnet,astarnet,adaprop,rgcn,kbgat} for KGs are \textit{composition-based}~\cite{compgcn} (i.e., messages are obtained with a binary operator using the message source embeddings and the relation type), thus directly benefit from this generic optimization. We extend RSPMM to support more complex composition operators, e.g., $2\times2$ block-diagonal matrix multiplication, Givens rotation and circular correlation. We extend the experiments of~\cite{mai2026rspmm} 
with comparisons against language model methods~\cite{simkgc,kermit} for KG representation learning. 

\section{Approach}

We assume $d\in \mathbb{N}$ to be an even integer. We refer the reader to~\cite{mai2026rspmm} for the definitions of $\cat$, $+$,$\odot $,$ \odotC$ , $\boxtimes$, $\mathrm{BlockDiag}$, $\mathrm{Rot}$, $\mathrm{Ref}$ and other notations. We focus on the sum aggregation ($\oplus= +$), which is highly popular in the GNN literature, and has desirable properties w.r.t. linearity and distributivity.
\paragraph{RSPMM.} Let $\phi : \R^{d_1} \times \R^{d_2} \to \R^{d_3}$, and let $\rspmm{\phi}{\cdot}{\cdot}{\cdot}$ (or $\RSPMM_\phi$) be defined as follows. For any adjacency tensor $A \in \R^{\abs{\calE} \times \abs{\calR} \times \abs{\calE}}$  of $\calG$, $\mathbf{H} \in \R^{\abs{\calE}\times d}$ and $\mathbf{Z} \in \R^{\abs{\calR}\times m}$,
\begin{equation}\label{def:rspmm}
    \forall i \in \calE, \left(\rspmm{\phi}{A}{\mathbf{H}}{ \mathbf{Z}}  \right)_i = \sum_{(j,r) \in \calN_{\mathrm{in}}(i)} A_{jri} \phi \left( \mathbf{H}_j,\mathbf{Z}_r \right).
\end{equation}

\begin{table}[t]
    \centering
    \caption{Extension of RSPMM in $\mathcal{O}(\abs{\calE}d)$ space complexity for operators inspired by shallow KGEs.}
    \label{tab:rspmm-extension}
    \resizebox{\textwidth}{!}{
    \begin{tabular}{llc}
    \toprule
    Definition of  $\phi(\bfh, \bfz)$ & Remarks \\
    \midrule
       $\bfh + \bfz$ (Addition) & Inspired by TransE~\cite{transe}, implemented by standard RSPMM~\cite{nbfnet} \\
       $\bfh \odot \bfz$ (Multiplication)& Inspired by DistMult~\cite{distmult}, implemented by standard RSPMM~\cite{nbfnet}\\
       $\bfh \odotC \bfz$ (Complex multiplication) & Inspired by ComplEx~\cite{complex} \\
       $\bfh \star \bfz$ (Circular correlation) & Inspired by HolE~\cite{nickel2015holographicembeddingsknowledgegraphs} \\
       $\bfh \cdot \mathrm{Rot}(\bfz)$ (Givens rotation) & Inspired by RotatE~\cite{rotate}. Givens reflection can also be implemented. \\
       $\mathbf{h} \boxtimes \mathbf{z} = \mathbf{h} \cdot \mathrm{BlockDiag}(\mathbf{z})$ (see~\cite{mai2026rspmm}) & Inspired by RESCAL~\cite{rescal}, with $2\times 2$ block-diagonal linear transformations \\
       $\phi\left(\bfh,\begin{bmatrix} \bfz_1 \cat \bfz_2\end{bmatrix}\right) = \phi_{0}\left(\bfh, \bfz_1\right) + \bfz_2$ & Assumes $d\geq \abs{\calR}$ and $\RSPMM_{\phi_0}$ can be implemented in $\mathcal{O}(\abs{\calE}d)$ space complexity \\
    \bottomrule
    \end{tabular}
    }
\end{table}

We propose implementations to compute $\RSPMM_\phi(A, \bfH, \bfZ)$ in $\mathcal{O}(\abs{\calE}d)$ space complexity, similarly to~\cite{nbfnet}, for a wider variety of operators $\phi$ shown in Table~\ref{tab:rspmm-extension}. See our code (\href{https://github.com/boschresearch/relational-message-passing}{https://github.com/boschresearch/relational-message-passing}) and examples below.

\paragraph{$2\times 2$ Block-diagonal matrix multiplication.} With $\bfH = [\mathbf{X} \cat \mathbf{Y} ] \in \mathbb{R}^{\abs{\calE}\times 2d}$ and $\bfZ = [\mathbf{P} \cat \mathbf{S} \cat \mathbf{Q} \cat \mathbf{R}] \in \mathbb{R}^{\abs{\calR}\times 4d}$, it can be proven that $\rspmm{\boxtimes}{A}{\bfH}{\bfZ} = \rspmm{\odot}{A}{\bfH}{[\mathbf{P} \cat \mathbf{S} ] } +\rspmm{\odot}{A}{ [\mathbf{Y} \cat \mathbf{X}]}{[\mathbf{Q} \cat \mathbf{R} ]}$. 
This provides a direct implementation of $\RSPMM_\boxtimes$ using two calls of $\RSPMM_\odot$, whose space complexity is $\mathcal{O}(\abs{\calE}d)$. 
$\RSPMM_\phi$ for complex pointwise multiplication and Givens rotation/reflection can be viewed as applications of $\RSPMM_\boxtimes$ for a well-designed $\mathbf{Z}$ (constructed from the original relation embeddings). $\RSPMM_\star$ can be computed with $\RSPMM_{\odotC}$ by exploiting the convolution theorem~\cite{oppenheim1999discrete}.

\paragraph{Additive Bias.} 
Assume that $\RSPMM_{\phi_0}$ is computable in $\mathcal{O}(\abs{\calE}d)$ for some operator $\phi_0$. We split $\mathbf{Z}$ into $\mathbf{Z}^{(1)}$ and $\mathbf{Z}^{(2)}$ vertically, s.t. $\rspmm{\phi_0}{A}{\mathbf{H}}{\mathbf{Z}^{(1)}}$ can be computed. We  then compute the matrix $B$, defined as $ B_{ir} = \sum_{j=1}^{\abs{\mathcal{E}}} A_{jri}$ of size $\abs{\mathcal{E}}\abs{\mathcal{R}} \leq \abs{\mathcal{E}}d$ when $d \geq \lvert \calR \rvert$. Then, it holds: 
$\rspmm{\phi}{A}{\bfH}{\bfZ} = \rspmm{\phi_0}{A}{\bfH}{\bfZ^{(1)}} + B\bfZ^{(2)}.$

\paragraph{Generalization.} In theory, it is possible to propose memory-efficient implementations for
\begin{equation}
   \mathrm{GRSPMM}_{\phi, \oplus} : (A, \bfH, \bfZ) \mapsto \bigoplus_{(j,r) \in \calN_{\mathrm{in}}(i)} \alpha_{jri} \phi(\bfH_j, \bfZ_r)
\end{equation}
for $\oplus \in \{+, \min, \max\}$ and $\phi$ in Table~\ref{tab:rspmm-extension}, which is more general. However, in case  $\oplus \in \{\min, \max\}$, new dedicated CUDA kernels would be needed, and these are out of scope for this work.

\paragraph{Theoretical Motivations.} Given entity embeddings $\bfH$ and relation embeddings $\bfZ$, Eq.~\ref{def:rspmm} can be viewed as computing the new representation $\left(\rspmm{\phi}{A}{\bfH}{\bfZ}\right)_i$ of $i \in \calE$ as a weighted sum of predictions of the embedding of $i$, relatively to the embeddings of its neighbors $j$ of relation $r$, given by $\phi(\bfH_j,\bfZ_r)$, so in a trained, vanilla composition-based GNN each layer already encodes the graph's structural information. These predictions are tied to $\phi$, which can be inspired from KGEs by design~\cite{compgcn}. However, in terms of relational patterns, different KG embeddings may better express different relational patterns~\cite{pavlović2023expressivespatiofunctionalembeddingknowledge}. For instance, TransE~\cite{transe} can easily model relational inverses (e.g., $(h,r_1,t) \in \calT$ if and only if $(t,r_2,h) \in \calT$, by using $\bfz_{r_2} = -\bfz_{r_1}$) and relational compositions ($(h,r_1,t_1), (t_1, r_2, t) \in \calT \iff (h,r,t) \in \calT$, with $\bfz_{r} = \bfz_{r_1} + \bfz_{r_2}$), but cannot model symmetric relations ($(h,r,t) \in \calT \iff (t,r,h) \in \calT$) well, whereas RotatE~\cite{rotate} can model all of the above. Therefore, the choice of $\phi$ is semantically meaningful for composition-based GNNs for KGs, and 
the ability to represent desirable patterns may lead to performance improvements. Our contribution bridges the gap between theoretical relational expressiveness and practical scalability challenges, by extending the catalogue of affordable $\phi$ functions.

\section{Experiments}

\paragraph{Setup.} We consider the following popular KG tasks: entity classification and KGC. We conduct multiple experiments to confirm the effectiveness and usefulness of our method: (a) re-implement the MP method of existing GNNs that directly benefit of our RSPMM extension (e.g.,~\cite{rgcn,compgcn,chen2021rgatrelationalgraphattention}), and compare the computational costs, (b) create lightweight \textit{prototypical models} inspired 
by existing, well-established architectures and compare them against SOTA methods,  (c) compare GNN methods against other popular language model (LM) embedding-based methods (e.g.,~\cite{simkgc,kermit}).  We refer the reader to~\cite{mai2026rspmm} for further details (e.g., training, evaluation, model categories and our code\footnote{\href{https://github.com/boschresearch/scalable-gnn-lp}{https://github.com/boschresearch/scalable-gnn-lp}}) for (a) and (b). Dataset statistics are shown in Table~\ref{tab:data-statistics}. Experiments are run with a NVIDIA A100 GPU, and an AMD Epyc 7543 @ 2.80 GHz CPU. We cut off experiments that take over 24 hours. In case of cutoff (OOT), the best previous checkpoint (over the validation set) is loaded to evaluate the model.

\begin{table}[t]
\centering
\caption{KG statistics. MCNC, resp. MLNC, stands for multiclass, resp. multilabel,  node classification.}
\label{tab:data-statistics}
\resizebox{0.8\textwidth}{!}{
\begin{tabular}{cccccccc}
\toprule
Dataset & Task & $\abs{\calE}$ & $\abs{\calR}$ & $\abs{\calT}$ & Train & Valid & Test \\
\midrule
WN18RR~\cite{dettmers2018convolutional2dknowledgegraph} & KGC & 40,943 & 11 & 86,835 & 86,835 & 3,034 & 3,134 \\
FB15k-237~\cite{toutanova-chen-2015-observed} & KGC & 14,541 & 237 & 272,115 & 272,115 & 17,535 & 20,466 \\
\hline
AM~\cite{ristoski2016collection} & MCNC & 1,666,764 & 265 & 5,988,321 & 802 & - & 198\\ 
AM+~\cite{bloem2021kgbench} & MCNC & 1,153,679 & 33 & 2,521,046 & 13,423 & 20,000 & 20,000 \\
dblp~\cite{bloem2021kgbench} & MCNC & 4,470,778 & 68 & 21,985,048 & 26,535 & 20,000 & 20,000 \\ 
\bottomrule
\end{tabular}
}
\end{table}

\paragraph{Prototypical Models.} We propose two prototypical models inspired by earlier works that benefit 
from an extended RSPMM: (1) MODEL-A, which modifies CompGCN~\cite{compgcn} as follows: (1a) the relation updates are carried out by a MLP instead of a linear layer, (1b) Entity and relation transformations are carried out \emph{before} MP; (2) MODEL-B, which modifies AdaProp~\cite{adaprop} primarily by (2a) using query-dependent initializations, and (2b) 
using a shared entity scorer for all layers. 
As KGs with textual descriptions are common, we extended experiments from~\cite{mai2026rspmm} with text, to 
show that GNNs with frozen LMs scale better than LM-only methods in these scenarios. We create initial entity features by embedding their descriptions with a frozen LM (all-mpnet-base-v2~\cite{reimers-2019-sentence-bert,song2020mpnetmaskedpermutedpretraining}), and feed them to our (R-MPNN) GNNs.

\begin{figure}[t]
    \centering
    \includegraphics[width=0.7\linewidth]{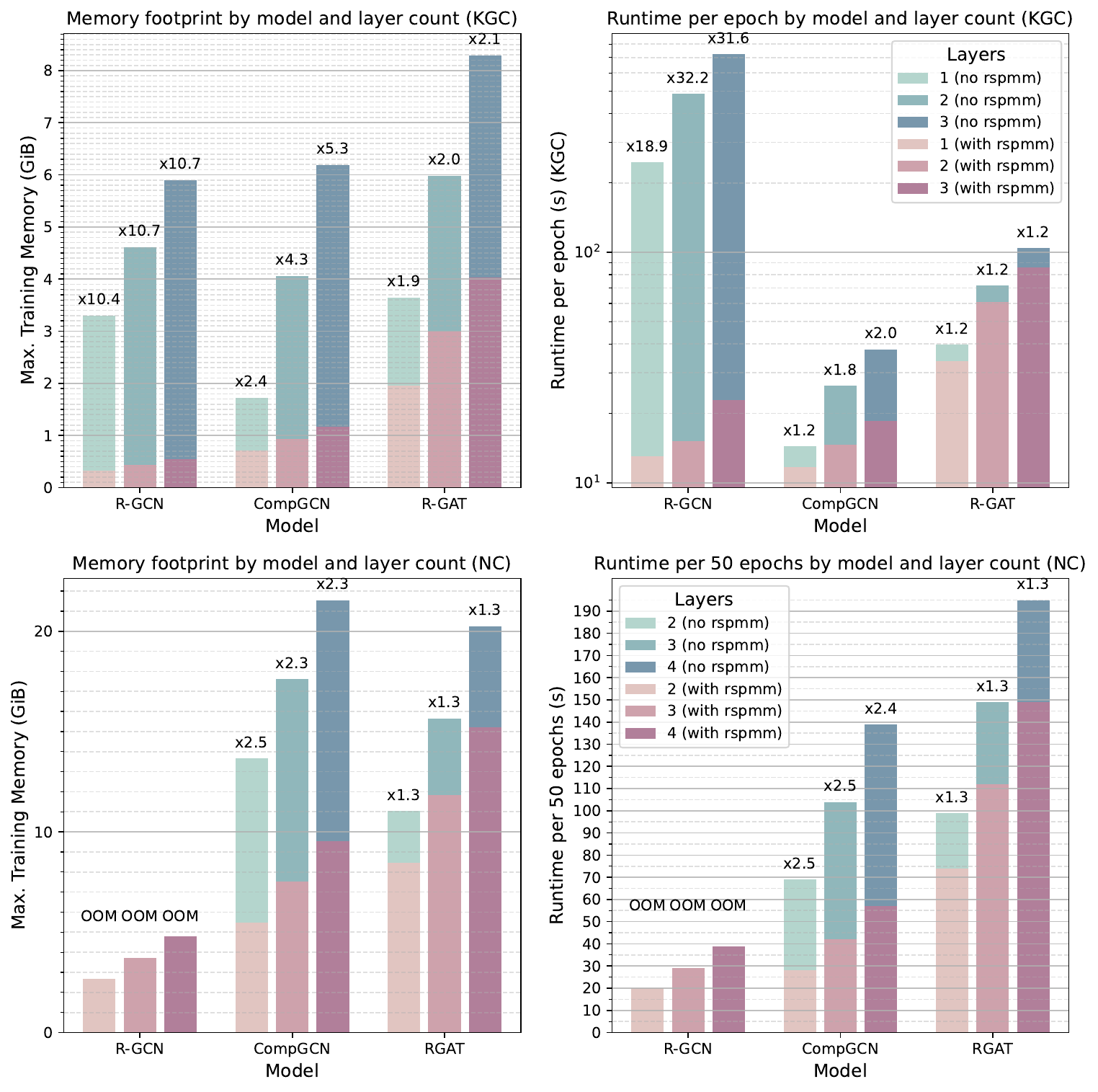}
    \caption{Runtimes and memory footprints during training for various GNN models. The top figures are results on FB15k-237~\cite{toutanova-chen-2015-observed}, and the bottom figures are results on AM~\cite{ristoski2016collection}. Legends apply to their respective rows.}
    \label{fig:scalability-results}
\end{figure}

\paragraph{Results.} Experiments with existing models (Figure~\ref{fig:scalability-results}, where CompGCN and R-GAT use $\phi=\star$) show that our RSPMM extension results in substantial scalability improvements without sacrificing performance (as the model remains theoretically equivalent), independently of model or task, up to $\times 10.7$ reductions in memory footprint and $\times 32.2$ reductions in runtime. This also holds for web-scale datasets (\texttt{dblp}~\cite{bloem2021kgbench} with 4.5M entities, 22M edges) where R-GCN uses up to 33.8 GiB during a full hyperparameter search (final accuracy: 0.72), while running out of memory without RSPMM. On KGC, Table~\ref{tab:kgc-results} shows that not only our implementations are substantially resource-friendlier than their counterparts within their respective category, they also perform very competitively. 
Furthermore, against LM-based methods, R-MPNNs with RSPMM with initial entity features perform comparably or better than methods involving the training of the LM, for substantially less memory and runtime.

\paragraph{Stress Tests.} We also carry out a fully fledged optimization pipeline with 10 trials of hyperparameter search on extremely large testbeds (AM+ and dblp), using R-GCN. We allow up to 72h of runtime. We use up to 4 layers and a hidden dimension up to 64. On AM+, the largest memory footprint occupied with our RSPMM-based implementation is \textbf{12.6 GiB over 11 hours} (including data preprocessing), versus \underline{21.5 GiB over 14 hours} without RSPMM (both having .92 final accuracy, which bests the reported performance in the paper). On DBLP, our implementation achieves a final accuracy of .72 with a maximal footprint of \textbf{33.8 GiB over 3 days}, while the base R-GCN \underline{runs out of memory}.

\begin{table}[t!]
    \centering
    \caption{Runtimes GPU memory footprints and performance metrics of KGC models on WN18RR and on FB15k-237. Overall best results are in bold, whereas other notable results (e.g. top results per category) are underlined.}
    \resizebox{\textwidth}{!}{
    \begin{tabular}{ccccccccccccc}
        \toprule
        
         \textbf{Category} & \textbf{Method} & \multicolumn{5}{c}{\textbf{WN18RR}\cite{dettmers2018convolutional2dknowledgegraph}} & & \multicolumn{5}{c}{\textbf{FB15k-237}~\cite{toutanova-chen-2015-observed}}\\
         \cline{3-7} \cline{9-13}
         & & Param. & Train T. & Train mem. & MRR & H@10 & & Param. & Train T. & Train mem. & MRR & H@10\\
         \midrule
          \multirow{2}*{KGE} & ComplEx-N3~\cite{complexN3}  & 40.975M & \underline{19 m} & \underline{\textbf{0.80 GiB}} & .469 & .541 & & 15.01M & \underline{31 m} & \underline{\textbf{0.30 GiB}} & \underline{.362} & \underline{.554}\\
         &RotH~\cite{lowdimensionalkge} & 20.50M& 65 m & 6.77 GiB & .491 & .579 & & 7.98M & 312 m & 6.47 GiB & .306 & .490\\
                  \hline
         \multirow{6}*{R-MPNN} & R-GCN (DGL)~\cite{rgcn} & 10.66M & 21 m & 2.46 GiB &  .389 & .429 & & 4.88M & \textbf{21 m} & 2.32 GiB & .263 & .457\\
         & CompGCN~\cite{compgcn} & 12.07M & 152 m & \underline{\textbf{0.85 GiB}} & .467 & .531 & & 9.45M& 110 m & 1.67 GiB & .349 & .529\\
         & RAGAT~\cite{ragat} & 68.24M & 1105 m & 3.07 GiB & .489 & .564 & & 17.77M & 425 m & 5.45 GiB & \underline{.364} & \underline{.548}\\
         & R-GCN (our impl.) & 10.93M & \underline{\textbf{5 m}}& \underline{1.49 GiB} & .486 & .596 & & 4.89M & \underline{30 m} & \underline{0.55 GiB} & .337 & .518 \\
         & $r$-GAT (our impl.) & 13.02M & \underline{16 m} & 2.50 GiB & \underline{.526} & \underline{.644} & & 7.19M & 121 m & 4.03 GiB & \underline{.365} & \underline{.554} \\
          & MODEL-A (ours) & 12.46M & \underline{\textbf{5 m}} & \underline{1.31 GiB} & \underline{.531} & \underline{.638 }& & 5.94M & \underline{32 m} & \underline{1.27 GiB} & \underline{.365} & \underline{.555}\\
          \hline
         \multirow{4}*{C-MPNN} & NBFNet~\cite{nbfnet} & 90K & 521 m & 26.12 GiB & .548 & \underline{.657} & & 3.10M &  1193 m & 18.99 GiB & \underline{\textbf{.416}} & \underline{\textbf{.598}}\\
         & A*Net~\cite{astarnet} & 90K & \underline{361 m }& 3.03 GiB & .539 & .644 & & 3.10M& OOT & 13.79 GiB & \underline{.413} & .586\\
         & AdaProp~\cite{adaprop} & 78K & 434 m & 8.95 GiB & .555 & \underline{.655} & & 195K & OOT & 14.94 GiB & .286 & .427\\
         & MODEL-B (ours) & 132K & 448 m & \underline{1.99 GiB} & \underline{.569} & \underline{.666} & & 7.07M & \underline{446 m} & \underline{2.79 GiB} & \underline{\textbf{.417}} &  \underline{\textbf{.596}} \\
         \midrule
       \multirow{4}*{LM} & SimKGC~\cite{simkgc} & 218M & 97 m & 61.08 GiB & \textbf{\underline{.673}} & .805 & & 218M & 86 m & 61.09 GiB & .337 & .512 \\ 
        & KERMIT~\cite{kermit} &  \multicolumn{5}{c}{Out of Memory (> 79.27 GiB)} & & \multicolumn{5}{c}{Out of Memory (> 79.27 GiB)} \\
        & R-GCN (our impl.) + Frozen LM & 713K & 11 m & \underline{1.62 GiB} & .586 & \underline{.808} & &  1.87M & 78 m & \underline{1.38 GiB} & .353 & .541 \\
        & MODEL-A + Frozen LM & 2.24M & 17 m & \underline{1.52 GiB} & \underline{.660} &  \underline{\textbf{.818}} & & 2.47M & 51 m & \underline{1.34 GiB} & \underline{.367} & \underline{.554} \\
        \bottomrule
    \end{tabular}
    }
    \label{tab:kgc-results}
\end{table}

\section{Conclusion}

We introduce an extension of RSPMM for diverse composition operators, readily applicable to many existing GNNs for KGs. Extensive experiments show that not only do relational GNNs benefit from significant lossless scalability gains by switching implementations, these improvements also allow one to afford better architectures with the same computational budget, independently of task or model, that can rival SOTA methods for substantially fewer resources. We view RSPMM as a useful tool for KG Machine Learning practitioners to speed up iteration cycles and tackle larger scale problems.

\section*{Declaration on Generative AI}
  The author(s) have not employed any Generative AI tools.

\bibliography{sample-ceur}

@String{Computing = "Computing" }

@String{Springer = "Springer-Verlag" }

@misc{R,
    title = {R: A Language and Environment for Statistical Computing},
    author = {{R Core Team}},
    organization = {R Foundation for Statistical Computing},
    address = {Vienna, Austria},
    year = {2019},
    url = {https://www.R-project.org/},
}

@misc{kepler,
      title={KEPLER: A Unified Model for Knowledge Embedding and Pre-trained Language Representation}, 
      author={Xiaozhi Wang and Tianyu Gao and Zhaocheng Zhu and Zhengyan Zhang and Zhiyuan Liu and Juanzi Li and Jian Tang},
      year={2020},
      eprint={1911.06136},
      archivePrefix={arXiv},
      primaryClass={cs.CL}, 
}

@misc{simkgc,
      title={SimKGC: Simple Contrastive Knowledge Graph Completion with Pre-trained Language Models}, 
      author={Liang Wang and Wei Zhao and Zhuoyu Wei and Jingming Liu},
      year={2022},
      eprint={2203.02167},
      archivePrefix={arXiv},
      primaryClass={cs.CL}, 
}

@misc{kermit,
      title={KERMIT: Knowledge Graph Completion of Enhanced Relation Modeling with Inverse Transformation}, 
      author={Haotian Li and Bin Yu and Yuliang Wei and Kai Wang and Richard Yi Da Xu and Bailing Wang},
      year={2024},
      eprint={2309.14770},
      archivePrefix={arXiv},
      primaryClass={cs.CL}, 
}

@inproceedings{transe,
author = {Bordes, Antoine and Usunier, Nicolas and Garcia-Dur\'{a}n, Alberto and Weston, Jason and Yakhnenko, Oksana},
title = {Translating embeddings for modeling multi-relational data},
year = {2013},
publisher = {Curran Associates Inc.},
address = {Red Hook, NY, USA},
booktitle = {Proceedings of the 26th International Conference on Neural Information Processing Systems - Volume 2},
pages = {2787–2795},
numpages = {9},
location = {Lake Tahoe, Nevada},
series = {NIPS'13}
}

@misc{distmult,
      title={Embedding Entities and Relations for Learning and Inference in Knowledge Bases}, 
      author={Bishan Yang and Wen-tau Yih and Xiaodong He and Jianfeng Gao and Li Deng},
      year={2015},
      eprint={1412.6575},
      archivePrefix={arXiv},
      primaryClass={cs.CL}, 
}

@misc{complex,
      title={Complex Embeddings for Simple Link Prediction}, 
      author={Théo Trouillon and Johannes Welbl and Sebastian Riedel and Éric Gaussier and Guillaume Bouchard},
      year={2016},
      eprint={1606.06357},
      archivePrefix={arXiv},
      primaryClass={cs.AI}, 
}

@misc{complexN3,
      title={Canonical Tensor Decomposition for Knowledge Base Completion}, 
      author={Timothée Lacroix and Nicolas Usunier and Guillaume Obozinski},
      year={2018},
      eprint={1806.07297},
      archivePrefix={arXiv},
      primaryClass={stat.ML}, 
}

@misc{rotate,
      title={RotatE: Knowledge Graph Embedding by Relational Rotation in Complex Space}, 
      author={Zhiqing Sun and Zhi-Hong Deng and Jian-Yun Nie and Jian Tang},
      year={2019},
      eprint={1902.10197},
      archivePrefix={arXiv},
      primaryClass={cs.LG}, 
}

@misc{lowdimensionalkge,
      title={Low-Dimensional Hyperbolic Knowledge Graph Embeddings}, 
      author={Ines Chami and Adva Wolf and Da-Cheng Juan and Frederic Sala and Sujith Ravi and Christopher Ré},
      year={2020},
      eprint={2005.00545},
      archivePrefix={arXiv},
      primaryClass={cs.LG}, 
}

@misc{dettmers2018convolutional2dknowledgegraph,
      title={Convolutional 2D Knowledge Graph Embeddings}, 
      author={Tim Dettmers and Pasquale Minervini and Pontus Stenetorp and Sebastian Riedel},
      year={2018},
      eprint={1707.01476},
      archivePrefix={arXiv},
      primaryClass={cs.LG}, 
}

@misc{nickel2015holographicembeddingsknowledgegraphs,
      title={Holographic Embeddings of Knowledge Graphs}, 
      author={Maximilian Nickel and Lorenzo Rosasco and Tomaso Poggio},
      year={2015},
      eprint={1510.04935},
      archivePrefix={arXiv},
      primaryClass={cs.AI}, 
}

@misc{pavlović2023expressivespatiofunctionalembeddingknowledge,
      title={ExpressivE: A Spatio-Functional Embedding For Knowledge Graph Completion}, 
      author={Aleksandar Pavlović and Emanuel Sallinger},
      year={2023},
      eprint={2206.04192},
      archivePrefix={arXiv},
      primaryClass={cs.LG},
      url={https://arxiv.org/abs/2206.04192}, 
}

@misc{rgcn,
      title={Modeling Relational Data with Graph Convolutional Networks}, 
      author={Michael Schlichtkrull and Thomas N. Kipf and Peter Bloem and Rianne van den Berg and Ivan Titov and Max Welling},
      year={2017},
      eprint={1703.06103},
      archivePrefix={arXiv},
      primaryClass={stat.ML}, 
}

@misc{compgcn,
      title={Composition-based Multi-Relational Graph Convolutional Networks}, 
      author={Shikhar Vashishth and Soumya Sanyal and Vikram Nitin and Partha Talukdar},
      year={2020},
      eprint={1911.03082},
      archivePrefix={arXiv},
      primaryClass={cs.LG}, 
}

@misc{kbgat,
      title={Learning Attention-based Embeddings for Relation Prediction in Knowledge Graphs}, 
      author={Deepak Nathani and Jatin Chauhan and Charu Sharma and Manohar Kaul},
      year={2019},
      eprint={1906.01195},
      archivePrefix={arXiv},
      primaryClass={cs.LG}, 
}

@misc{chen2021rgatrelationalgraphattention,
      title={r-GAT: Relational Graph Attention Network for Multi-Relational Graphs}, 
      author={Meiqi Chen and Yuan Zhang and Xiaoyu Kou and Yuntao Li and Yan Zhang},
      year={2021},
      eprint={2109.05922},
      archivePrefix={arXiv},
      primaryClass={cs.AI}, 
}

@ARTICLE{ragat,
  author={Liu, Xiyang and Tan, Huobin and Chen, Qinghong and Lin, Guangyan},
  journal={IEEE Access}, 
  title={RAGAT: Relation Aware Graph Attention Network for Knowledge Graph Completion}, 
  year={2021},
  volume={9},
  number={},
  pages={20840-20849},
  doi={10.1109/ACCESS.2021.3055529}}

@misc{kracl,
      title={KRACL: Contrastive Learning with Graph Context Modeling for Sparse Knowledge Graph Completion}, 
      author={Zhaoxuan Tan and Zilong Chen and Shangbin Feng and Qingyue Zhang and Qinghua Zheng and Jundong Li and Minnan Luo},
      year={2023},
      eprint={2208.07622},
      archivePrefix={arXiv},
      primaryClass={cs.AI}, 
}

@misc{nbfnet,
      title={Neural Bellman-Ford Networks: A General Graph Neural Network Framework for Link Prediction}, 
      author={Zhaocheng Zhu and Zuobai Zhang and Louis-Pascal Xhonneux and Jian Tang},
      year={2022},
      eprint={2106.06935},
      archivePrefix={arXiv},
      primaryClass={cs.LG}, 
}

@misc{astarnet,
      title={A*Net: A Scalable Path-based Reasoning Approach for Knowledge Graphs}, 
      author={Zhaocheng Zhu and Xinyu Yuan and Mikhail Galkin and Sophie Xhonneux and Ming Zhang and Maxime Gazeau and Jian Tang},
      year={2023},
      eprint={2206.04798},
      archivePrefix={arXiv},
      primaryClass={cs.AI}, 
}

@misc{adaprop,
      title={AdaProp: Learning Adaptive Propagation for Graph Neural Network based Knowledge Graph Reasoning}, 
      author={Yongqi Zhang and Zhanke Zhou and Quanming Yao and Xiaowen Chu and Bo Han},
      year={2023},
      eprint={2205.15319},
      archivePrefix={arXiv},
      primaryClass={cs.LG}, 
}

@misc{feeney2021relationmatterssamplingscalable,
      title={Relation Matters in Sampling: A Scalable Multi-Relational Graph Neural Network for Drug-Drug Interaction Prediction}, 
      author={Arthur Feeney and Rishabh Gupta and Veronika Thost and Rico Angell and Gayathri Chandu and Yash Adhikari and Tengfei Ma},
      year={2021},
      eprint={2105.13975},
      archivePrefix={arXiv},
      primaryClass={cs.LG}, 
}

@inproceedings{reimers-2019-sentence-bert,
  title = "Sentence-BERT: Sentence Embeddings using Siamese BERT-Networks",
  author = "Reimers, Nils and Gurevych, Iryna",
  booktitle = "Proceedings of the 2019 Conference on Empirical Methods in Natural Language Processing",
  month = "11",
  year = "2019",
  publisher = "Association for Computational Linguistics",
}

@misc{song2020mpnetmaskedpermutedpretraining,
      title={MPNet: Masked and Permuted Pre-training for Language Understanding}, 
      author={Kaitao Song and Xu Tan and Tao Qin and Jianfeng Lu and Tie-Yan Liu},
      year={2020},
      eprint={2004.09297},
      archivePrefix={arXiv},
      primaryClass={cs.CL},
      url={https://arxiv.org/abs/2004.09297}, 
}

@article{wikidata,
author = {Vrande\v{c}i\'{c}, Denny and Kr\"{o}tzsch, Markus},
title = {Wikidata: a free collaborative knowledgebase},
year = {2014},
issue_date = {October 2014},
publisher = {Association for Computing Machinery},
address = {New York, NY, USA},
volume = {57},
number = {10},
issn = {0001-0782},
doi = {10.1145/2629489},
journal = {Commun. ACM},
month = sep,
pages = {78–85},
numpages = {8}
}

@inproceedings{toutanova-chen-2015-observed,
  author       = {Kristina Toutanova and
                  Danqi Chen},
  title        = {Observed versus latent features for knowledge base and text inference},
  booktitle    = {Proceedings of the 3rd Workshop on Continuous Vector Space Models
                  and their Compositionality, {CVSC} 2015, Beijing, China, July 26-31,
                  2015},
  pages        = {57--66},
  publisher    = {Association for Computational Linguistics},
  year         = {2015},
    doi          = {10.18653/V1/W15-4007},
}

@inproceedings{Mahdisoltani2015YAGO3AK,
  title={YAGO3: A Knowledge Base from Multilingual Wikipedias},
  author={Farzaneh Mahdisoltani and Joanna Asia Biega and Fabian M. Suchanek},
  booktitle={Conference on Innovative Data Systems Research},
  year={2015},
  url={https://api.semanticscholar.org/CorpusID:6611164}
}

@inproceedings{ristoski2016collection,
  title={A collection of benchmark datasets for systematic evaluations of machine learning on the semantic web},
  author={Ristoski, Petar and De Vries, Gerben Klaas Dirk and Paulheim, Heiko},
  booktitle={International semantic web conference},
  pages={186--194},
  year={2016},
  organization={Springer}
}

@article{liang2023hyper,
  title={Hyper-node relational graph attention network for multi-modal knowledge graph completion},
  author={Liang, Shuang and Zhu, Anjie and Zhang, Jiasheng and Shao, Jie},
  journal={ACM Transactions on Multimedia Computing, Communications and Applications},
  volume={19},
  number={2},
  pages={1--21},
  year={2023},
  publisher={ACM New York, NY}
}

@misc{ge-spmm,
      title={GE-SpMM: General-purpose Sparse Matrix-Matrix Multiplication on GPUs for Graph Neural Networks}, 
      author={Guyue Huang and Guohao Dai and Yu Wang and Huazhong Yang},
      year={2020},
      eprint={2007.03179},
      archivePrefix={arXiv},
      primaryClass={cs.DC}, 
}

@inproceedings{pyg,
  title={Fast Graph Representation Learning with {PyTorch Geometric}},
  author={Fey, Matthias and Lenssen, Jan E.},
  booktitle={ICLR Workshop on Representation Learning on Graphs and Manifolds},
  year={2019},
}

@inproceedings{rescal,
author = {Nickel, Maximilian and Tresp, Volker and Kriegel, Hans-Peter},
title = {A three-way model for collective learning on multi-relational data},
year = {2011},
isbn = {9781450306195},
publisher = {Omnipress},
address = {Madison, WI, USA},
booktitle = {Proceedings of the 28th International Conference on International Conference on Machine Learning},
pages = {809–816},
numpages = {8},
location = {Bellevue, Washington, USA},
series = {ICML'11}
}

@inproceedings{zhu2019graphvite,
    title={GraphVite: A High-Performance CPU-GPU Hybrid System for Node Embedding},
     author={Zhu, Zhaocheng and Xu, Shizhen and Qu, Meng and Tang, Jian},
     booktitle={The World Wide Web Conference},
     pages={2494--2504},
     year={2019},
     organization={ACM}
 }

@inproceedings{DGL-KE,
author = {Zheng, Da and Song, Xiang and Ma, Chao et al.
},
title = {DGL-KE: Training Knowledge Graph Embeddings at Scale},
year = {2020},
publisher = {Association for Computing Machinery},
address = {New York, NY, USA},
booktitle = {Proceedings of the 43rd International ACM SIGIR Conference on Research and Development in Information Retrieval},
pages = {739–748},
numpages = {10},
series = {SIGIR '20}
}

@misc{hamilton2018inductiverepresentationlearninglarge,
      title={Inductive Representation Learning on Large Graphs}, 
      author={William L. Hamilton and Rex Ying and Jure Leskovec},
      year={2018},
      eprint={1706.02216},
      archivePrefix={arXiv},
      primaryClass={cs.SI}, 
}

@article{relational-ml,
   title={A Review of Relational Machine Learning for Knowledge Graphs},
   volume={104},
   ISSN={1558-2256},
      DOI={10.1109/jproc.2015.2483592},
   number={1},
   journal={Proceedings of the IEEE},
   publisher={Institute of Electrical and Electronics Engineers (IEEE)},
   author={Nickel, Maximilian and Murphy, Kevin and Tresp, Volker and Gabrilovich, Evgeniy},
   year={2016},
   month=jan, pages={11–33} }

@misc{pytorch-biggraph,
      title={PyTorch-BigGraph: A Large-scale Graph Embedding System}, 
      author={Adam Lerer and Ledell Wu and Jiajun Shen and Timothee Lacroix and Luca Wehrstedt and Abhijit Bose and Alex Peysakhovich},
      year={2019},
      eprint={1903.12287},
      archivePrefix={arXiv},
      primaryClass={cs.LG}, 
}

@misc{nodepiece,
      title={NodePiece: Compositional and Parameter-Efficient Representations of Large Knowledge Graphs}, 
      author={Mikhail Galkin and Etienne Denis and Jiapeng Wu and William L. Hamilton},
      year={2022},
      eprint={2106.12144},
      archivePrefix={arXiv},
      primaryClass={cs.CL}, 
}

@inproceedings{recpiece,
  author       = {Ke Liang and
                  Yue Liu and
                  Hao Li and
                  Lingyuan Meng and
                  Suyuan Liu and
                  Siwei Wang and
                  Sihang Zhou and
                  Xinwang Liu},
  title        = {Clustering then Propagation: Select Better Anchors for Knowledge Graph
                  Embedding},
  booktitle    = {Advances in Neural Information Processing Systems 38: Annual Conference
                  on Neural Information Processing Systems 2024, NeurIPS 2024, Vancouver,
                  BC, Canada, December 10 - 15, 2024},
  year         = {2024}
}

@misc{stargraph,
      title={StarGraph: Knowledge Representation Learning based on Incomplete Two-hop Subgraph}, 
      author={Hongzhu Li and Xiangrui Gao and Linhui Feng and Yafeng Deng and Yuhui Yin},
      year={2023},
      eprint={2205.14209},
      archivePrefix={arXiv},
      primaryClass={cs.CL}, 
}

@InProceedings{rdf2vec,
author="Ristoski, Petar
and Paulheim, Heiko",
title="RDF2Vec: RDF Graph Embeddings for Data Mining",
booktitle="The Semantic Web -- ISWC 2016",
year="2016",
publisher="Springer International Publishing",
address="Cham",
pages="498--514",
isbn="978-3-319-46523-4"
}

@article{wl-rdf,
title = {Substructure counting graph kernels for machine learning from RDF data},
journal = {Journal of Web Semantics},
volume = {35},
pages = {71-84},
year = {2015},
note = {Machine Learning and Data Mining for the Semantic Web (MLDMSW)},
issn = {1570-8268},
doi = {https://doi.org/10.1016/j.websem.2015.08.002},
author = {Gerben Klaas Dirk {de Vries} and Steven {de Rooij}}
}

@inproceedings{zhu2017iterative,
  title={Iterative Entity Alignment via Joint Knowledge Embeddings.},
  author={Zhu, Hao and Xie, Ruobing and Liu, Zhiyuan and Sun, Maosong},
  booktitle={IJCAI},
  volume={17},
  number={2017},
  pages={4258--4264},
  year={2017}
}

@misc{ge2021largeeaaligningentitieslargescale,
      title={LargeEA: Aligning Entities for Large-scale Knowledge Graphs}, 
      author={Congcong Ge and Xiaoze Liu and Lu Chen and Baihua Zheng and Yunjun Gao},
      year={2021},
      eprint={2108.05211},
      archivePrefix={arXiv},
      primaryClass={cs.DB},
      doi={https://doi.org/10.14778/3489496.3489504}, 
}

@inproceedings{
    bloem2021kgbench,
    title={kgbench: A Collection of Knowledge Graph Datasets for Evaluating Relational and Multimodal Machine Learning},
    author={Peter Bloem and Xander Wilcke and Lucas van Berkel and Victor de Boer},
    booktitle={Eighteenth Extended Semantic Web Conference - Resources Track},
    year={2021}
}

@inproceedings{mai2026rspmm,
  title={Knowledge Graph Representation Learning with Efficient Message Passing},
  author={Mai, Huu Tan and Chu, Cuong Xuan and Paulheim, Heiko and Stepanova, Daria},
  booktitle={International Semantic Web conference},
  year={2026},
}

@book{oppenheim1999discrete,
  title={Discrete-time signal processing},
  author={Oppenheim, Alan V},
  year={1999},
  publisher={Pearson Education India}
}




\end{document}